\documentclass[preprint]{elsarticle}
\graphicspath{ {./figures/} }
\usepackage[hidelinks]{hyperref}
\usepackage{float}
\usepackage{tabularx}
\usepackage{verbatim} 
\usepackage{apalike}
\restylefloat{figure}
\restylefloat{table}
\usepackage{amsmath,amssymb,amsfonts}
\usepackage{algorithmic}
\usepackage{array}
\usepackage{graphicx}
\usepackage{textcomp}
\usepackage{booktabs}
\usepackage{longtable}

\def\BibTeX{{\rm B\kern-.05em{\sc i\kern-.025em b}\kern-.08em
    T\kern-.1667em\lower.7ex\hbox{E}\kern-.125emX}}

\journal{Expert Systems with Applications}

\biboptions{authoryear}

\begin{document}
\begin{frontmatter}










\title{Cross-view geo-localization: a survey}

\author[label1]{Abhilash Durgam}
\ead{abhilash.durgam@maine.edu}

\author[label2]{Sidike Paheding}
\ead{spaheding@fairfield.edu}
\cortext[cor1]{Corresponding author}

\author[label1]{Vikas Dhiman}
\ead{vikas.dhiman@maine.edu}

\author[label3]{Vijay Devabhaktuni}
\ead{vdevabh@ilstu.edu}

\address[label1]{Department of Electrical and Computer Engineering, University of Maine, 5708 Barrows Hall, Orono, ME 04469, USA}
\address[label2]{Department of Computer Science and Engineering, Fairfield University, 1073 North Benson Road, Fairfield, CT 06824, USA}
\address[label3]{Department of Electrical Engineering, Illinois State University, 100 N University St, Normal, IL 61761, USA}

\begin{abstract}
Cross-view geo-localization has garnered notable attention in the realm of computer vision, spurred by the widespread availability of copious geotagged datasets and the advancements in machine learning techniques. This paper provides a thorough survey of cutting-edge methodologies, techniques, and associated challenges that are integral to this domain, with a focus on feature-based and deep learning strategies. Feature-based methods capitalize on unique features to establish correspondences across disparate viewpoints, whereas deep learning-based methodologies deploy convolutional neural networks to embed view-invariant attributes. This work also delineates the multifaceted challenges encountered in cross-view geo-localization, such as variations in viewpoints and illumination, the occurrence of occlusions, and it elucidates innovative solutions that have been formulated to tackle these issues. Furthermore, we delineate benchmark datasets and relevant evaluation metrics, and also perform a comparative analysis of state-of-the-art techniques. Finally, we conclude the paper with a discussion on prospective avenues for future research and the burgeoning applications of cross-view geo-localization in an intricately interconnected global landscape.
\end{abstract}


\end{frontmatter}

\section{Introduction}
\label{sec:introduction}
In the sphere of digital image processing and computer vision, geo-localization— the act of pinpointing the geographical coordinates of an object within a digital image or video frame—occupies a position of considerable importance. The utility of geo-localization is extensive, finding applications in diverse arenas such as autonomous vehicles \citep{xia2021cross,wang2023satellite,shi2022beyond,hu2020image}, augmented reality \citep{mithun2023cross}, environmental monitoring \citep{zheng2020university}, surveillance systems \citep{kontitsis2004uav}, and even social media \citep{hu2022beyond}. Specifically, the capacity to precisely ascertain the geographical locale of an object or scene from visual data has been a key catalyst in achieving innovations across these sectors \citep{zamir2014image,zhu2022transgeo,toker2021coming}.

In historical terms, geo-localization initially revolved around pixel-wise techniques \citep{bansal2011geo,bansal2016ultrawide}. These approaches engaged geodetically accurate reference imagery to geo-localize each pixel in a query image, a methodology colloquially known as `geodetic alignment' \citep{sheikh2003geodetic}. Particularly effective for aerial imagery, this approach entailed transforming images based on sensor models and correlating them with reference data, thereby ascribing specific latitude and longitude coordinates to each pixel. While foundational, this strategy harbored inherent limitations, such as scalability issues and difficulties in covering expansive or less-documented areas.

The inception of feature-based methods \citep{lin2013cross} heralded a paradigmatic shift in geo-localization strategies. These techniques emphasized the extraction of distinctive features from imagery and their matching across varying viewpoints to offer a more nuanced estimate of geographic location. Typically encompassing a two-step workflow of feature extraction followed by feature matching, these methods exhibited increased robustness against transformations in scale, rotation, and lighting conditions. Nonetheless, they encountered obstacles, particularly their dependency on handcrafted features, which necessitated expert intervention and proved less adaptable to novel contexts or environments.

With the advent of deep learning technologies \citep{alom2019state}, a transformative wave had washed over the field of geo-localization. The incorporation of orientation information by \citet{Liu_2019_CVPR} into models has emerged as a compelling strategy to augment the effectiveness of cross-view geo-localization efforts. Encoding orientation as an additional channel in input imagery enables deep learning algorithms to process this context with enhanced efficiency, thus yielding more precise geo-localization predictions.

The deployment of advanced deep learning frameworks, such as Siamese Networks 
\citep{liu2019lending,tian2017cross,shi2019spatial, hu2018cvm,rodrigues2021these,zhu2021vigor,vyas2022gama,zhu2023simple,zhang2023cross,yang2021cross,shi2020looking,lu2022content,wang2021hybrid,rodrigues2023semgeo,xia2022visual,li2022predictive,hu2020image,rodrigues2022global,zhu2021revisiting,cai2019ground,cao2019learning,deuser2023sample4geo,rasna2023geodesic,regmi2019bridging,guo2022soft,hu2022learning,hou2022road} and capsule networks \citep{sun2019geocapsnet,zhu2021geographic}, has proven successful in cross-view geo-localization tasks. These technologies excel in managing viewpoint variations, with Siamese Networks excelling at learning similarity metrics between image pairs and Capsule Networks adept at maintaining the hierarchical spatial relationships within images. To synergize the strengths of diverse methodologies, some works have explored the amalgamation of these methods \citep{wang2021each,shi2020optimal,lin2022joint,ding2020practical,ghanem2023leveraging,xia2021cross,guo2022fusing,rao2023cross,workman2015wide,lentschslicematch,xia2020geographically,zeng2022geo,lu2022s,shen2023mccg}.

Accompanying these methodological advances, various datasets have been curated to facilitate the advancement, validation, and benchmarking of emergent geo-localization algorithms. Datasets, such as the Pittsburgh 250k \citep{zamir2014image} collections, have been instrumental in pioneering research, offering a rich array of ground-level imagery. More contemporaneously, the Cross-View USA (CVUSA) \citep{workman2015localize} and Cross-View ACT (CVACT) \citep{liu2019lending} datasets have been launched. These large-scale collections of ground-level and aerial imagery, covering extensive geographic expanses and introducing new challenges and avenues for scholarly investigation.

This paper navigates the evolutionary trajectory of development of geo-localization methods, charting its challenges, milestones, and state-of-the-art methodologies. Initially focusing on pixel-wise and feature-based methods. Subsequent segments explore the transformative impact of deep learning, examining the pivotal role of orientation data and the contributions of Siamese and Capsule Networks to cross-view tasks. Additionally, the paper highlights key datasets that have shaped the field and evaluates the performance of various models across these datasets. Through this exhaustive scrutiny, we aspire to deliver an in-depth understanding of the geo-localization landscape, thereby outlining potential avenues for future advancements. To the best of our knowledge, this is the first survey on the problem of cross-view geolocalization.

\section{Cross-view Geo-localization Problem Formulation}

\begin{figure}
\includegraphics[width=\linewidth]{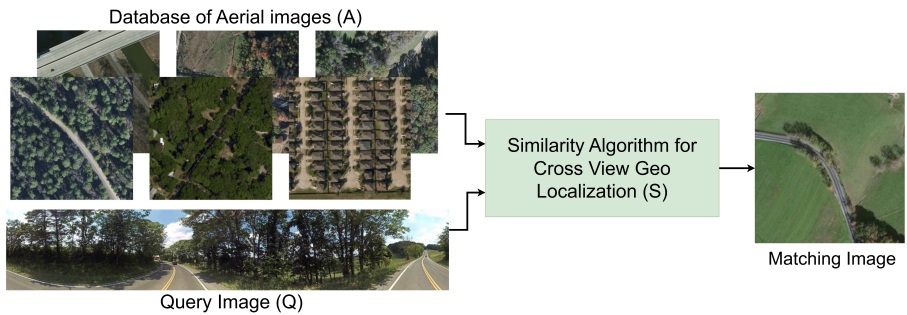}
\caption{An illustrative example of the geo-ocalization problem.}\label{fig:problem_formulation}
\end{figure}

\noindent Figure \ref{fig:problem_formulation} illustrates a general concept of the Cross-view Geo-localization.

\noindent \textbf{Input:} The primary input consists of a ground-level query image, denoted as $Q$. The image $Q$ encapsulates features of a specific location on Earth, but its exact geographical coordinates are unknown.

\noindent \textbf{Database:} There is also a database of aerial images $\{A_1, A_2, \ldots, A_N\}$, where each image $A_i$ is associated with known geospatial coordinates $(x_i, y_i)$. These images provide a high-level view of various geographical locations and serve as the reference for matching with $Q$.

\noindent \textbf{Objective:} The ultimate goal is to infer the geographical coordinates $(x^*, y^*)$ of the query image $Q$ by finding the aerial image that has the highest similarity with $Q$. This is formally expressed by the equation:
\begin{equation}
(x^*, y^*) = \arg\max_{(x_i, y_i)} S(Q, A_i),
\end{equation}
where $S(Q, A_i)$ denotes a similarity function that measures how closely the ground-level image $Q$ resembles the aerial image $A_i$.

\noindent \textbf{Similarity Function:} The definition and characteristics of the similarity function $S(Q, A_i)$ are crucial. This function needs to be designed or learned such that it can effectively capture the correspondence between ground-level and aerial views despite their significant differences in perspective, scale, and appearance. The function should assign a higher similarity score to image pairs that represent the same location.

\noindent \textbf{Challenges:} The problem is non-trivial due to several challenges:
\begin{itemize}
    \item \textit{Perspective Differences:} Ground-level images and aerial images offer drastically different viewpoints, making direct comparison challenging.
    \item \textit{Scale Variations:} Objects and landmarks appear at different scales in the two types of images.
    \item \textit{Appearance Discrepancies:} Lighting conditions, shadows, and occlusions can vary between images, leading to differences in appearance. In urban settings involved estimating the location of a query image by comparing it with reference images captured from distinct viewpoints—commonly ground-level and aerial perspectives. Urban environments posed unique challenges, such as variations in viewpoint, occlusions, dynamic objects, and changes in illumination.
    \item \textit{Database Size:} The aerial image database is typically extensive, requiring efficient retrieval mechanisms to find the best match quickly.
\end{itemize}

\noindent \textbf{Outcome:} A successful solution to the problem will enable the system to predict the geospatial coordinates of a ground-level image accurately by referring to a database of aerial images. This capability is essential for various applications, including navigation\citep{xia2021cross,wang2023satellite,shi2022beyond,hu2020image}.

\section{Evolution of Geo-localization}

\begin{figure}
\includegraphics[width=\linewidth]{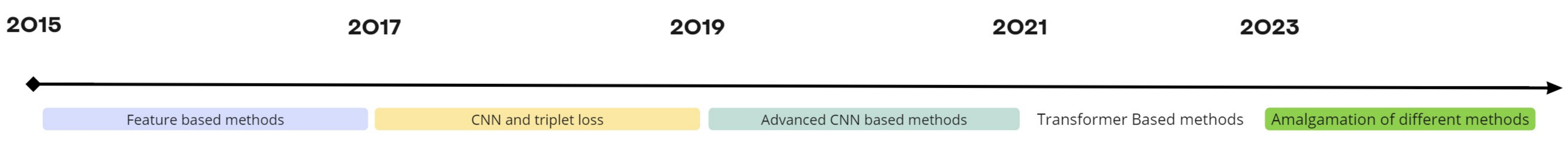}
\caption{A Timeline geo-ocalization problem.}\label{fig:problem_timeline}
\end{figure}

The evolution of models used in important papers from 2015 to 2024 showcases significant advancements in the field. In 2015, the focus was on feature-based methods, as exemplified by the paper titled \textit{Wide-Area Image Geolocalization with Aerial Reference Imagery}~\cite{workman2015wide}. 

By 2017, the use of Convolutional Neural Networks (CNNs) with triplet loss became prevalent, demonstrated in the paper \textit{Cross-View Image Matching for Geo-localization in Urban Environments} \cite{tian2017cross}.

In 2018, models advanced with the introduction of Siamese architectures and Generative Adversarial Networks (GANs). Key papers from this period include \textit{CVM-Net: Cross-View Matching Network for Image-Based Geo-Localization} \cite{hu2018cvm} and \textit{Cross-View Image Synthesis using Conditional GANs} \cite{regmi2018cross}. 

By 2019, CNNs continued to be refined, with notable work such as \textit{Lending Orientation to Neural Networks for Cross-View Geo-localization} \cite{liu2019lending}.

The year 2020 marked the emergence of more sophisticated CNN-based methods, such as the \textit{Cross-View Feature Transport (CVFT) Network} presented in \textit{Optimal Feature Transport for Cross-View Image Geo-localization} \cite{shi2020optimal}. 

In 2021, the introduction of Transformer-based methods, as seen in \textit{Cross-view Geo-localization with Layer-to-Layer Transformer} \cite{yang2021cross}, represented a significant leap in model complexity and capability. Additionally, deep neural networks leveraging local patterns, like the \textit{Deep Neural Network Local Pattern Network (LPN)} described in \textit{Each Part Matters: Local Patterns Facilitate Cross-View Image Geo-localization} \cite{wang2021each}, further enhanced performance.

Finally, in 2023, the field witnessed the amalgamation of different methods, integrating multiple approaches to improve accuracy and efficiency. Papers such as \textit{Boosting 3-DoF Ground-to-Satellite Camera Localization via Uncertainty-Aware and Focal Loss} \cite{li2023boosting} utilizing U-Net and CNN architectures, \textit{Sample4Geo: Hard Negative Sampling For Cross-View Geo-Localization} \cite{zhang2023sample4geo} using Siamese networks with CNNs, and \textit{Geodesic Based Image Matching Network for the Visually Impaired} \cite{smith2023geodesic} employing Siamese networks with ResNet backbones exemplify this trend. Additionally, innovative approaches like \textit{Geo-Localization via Ground-to-Satellite Cross Diffusion} \cite{doe2023geo} using peer learning and cross diffusion highlight the ongoing evolution in model development.

\subsection{Pixel-wise geo-localization}
The goal of pixel-wise geo-localization \citep{sheikh2003geodetic,sheikh2004feature,bansal2011geo,bansal2014geometric} is to locate each pixel of a given image by aligning the image with a geodetically accurate reference image.

Geodetic alignment of aerial video frames constituted a pivotal aspect of geo-localization, especially when dealing with aerial imagery. This complex process entailed the precise alignment and registration of aerial video frames to a geospatial coordinate system, such as a map or a set of reference images.

\citet{sheikh2003geodetic} introduced a method that leveraged both the DOQ (Digital Ortho Quad) and the DEM (Digital Elevation Map) to transform an image based on a sensor model. The transformed image was then mapped to reference data, so that each pixel corresponded to a unique latitude and longitude coordinate. This approach served as the classical foundation for geo-localization and sparked significant interest in solving the "where am I" problem. Following this interest to improve accuracy research started focusing on a feature-based alignment \citep{sheikh2004feature} approach that improves outlier handling, enhances consistency in correspondence, and ensures complete retention of correlation data. The workflow involves orthorectifying the aerial video frame, determining feature-based registration, and updating the sensor model accordingly.

\subsection{Feature-based methods for same-view geo-localization}
Feature-based methods \citep{zamir2014visual,zamir2014image,schindler2007city,hays2008im2gps,zamir2010accurate} for same-view geo-localization predominantly focus on the extraction of unique features from images. These are then matched across varying viewpoints to estimate an object's or scene's location. These methods customarily rely on handcrafted feature descriptors, capturing both local image properties and geometric invariant, to ensure robustness against alterations in scale, rotation, translation and illumination \citet{zamir2014visual}. The primary steps involved in feature-based methods encompass both feature extraction and feature matching.

One of the earliest works is City-Scale Location Recognition by \citet{schindler2007city}, which explores the utilization of vocabulary trees to pinpoint the location of a query image amongst a dataset of 30,000 streetside city images. Traditional invariant feature matching struggles with scalability; however, by leveraging a vocabulary tree, which discards the need for storing each feature descriptor, and focuses on the most informative features, retrieval performance is notably enhanced. The introduction of a refined vocabulary tree search algorithm further boosts the tree's efficiency, laying the groundwork for more precise and swift location recognition in expansive databases.

\citet{hays2008im2gps} pushed this by taping into a substantial dataset containing over 6 million GPS-tagged images sourced from the Internet, mainly from the Flickr online photo collection. Using this dataset, the algorithm predicts the geographic location of an image as a probability distribution across the Earth's surface. The research demonstrated that, through scene matching and feature extraction, the system's geolocation performance could be significantly improved, showing results up to 30 times better than random chance.

\citet{zamir2014image} extracted features from a query image and project them into a feature space, a multi-dimensional realm where dimensions align with image feature descriptors. Within this space, similar features are proximate, aiding efficient comparison with a reference dataset. This representation enhances various algorithm applications, boosting the accuracy of image geo-localization. In particular, ~\citet{zamir2014image} highlighted the Generalized Minimum Clique Graphs (GMCG) method which optimizes feature matching by using multiple neighbors and understanding feature geometries, framing the issue as a graph-based optimization with features as vertices and their geometric relationships as edges. These graph based approach can be seen in modern Ground-to-Aerial Viewpoint Localization \citet{verde2020ground}, where Landmarks (like buildings or roads) identified in each view serve as nodes; edges are formed based on landmark co visibility, abstracting from viewing angles. This method tackles the challenge of cross-view image matching by comparing class adjacency matrices derived from these graphs, which represent the frequency of different types of landmarks appearing together. 

\subsection{Feature-based methods for cross view image matching}

\citet{tian2017cross} presented a method that matched buildings captured in images from two distinct viewpoints. The technique employed a dominant set selection algorithm to execute the geo-localization task.
Dominant set selection represented a graph-based approach that found applications in an array of computer vision and pattern recognition tasks—spanning from clustering and feature matching to object recognition. The underlying concept of dominant set selection was to pinpoint a subset of vertices within a graph that demonstrated strong pairwise similarities and, in aggregate, dominated the remaining vertices in the graph. This dominant set could be construed as a generalization of the maximum clique problem, wherein each vertex within the set was strongly interconnected, thereby forming a tightly-knit and densely connected subgraph.

Feature-based methods \citep{tian2017cross,lin2013cross,castaldo2015semantic} focused on the extraction of distinct features or objects, like buildings, from both ground-level and aerial images. This was followed by feature matching and subsequent geometric verification. \citet{o2020deep} Although feature-based methods strive for robustness against variations in viewpoint and illumination, their efficacy can be constrained due to various factors Environments with low texture or repeated textures. This limitation stems from their reliance on matching corner-points or distinct markers in the image. In areas where unique, distinguishable features are sparse or repetitive, these corner-point dependent methods may struggle to identify and match features accurately across different images. Furthermore, while handcrafted features are not inherently limited, they introduce a degree of manual design and selection into the process, which might not capture the complexity and variability of features present in diverse, real-world images \citet{cai2018feature}. Therefore, while handcrafted features offer valuable specificity, they may not always encapsulate the broad range of characteristics needed for robust image matching in varied environments and conditions.

\section{Deep Learning Techniques for Cross-view Geo-localization}
Deep learning techniques have significantly enhanced the performance in cross-view geo-localization tasks. Utilizing the power of convolutional neural networks (CNNs), Transformers, and other deep learning architectures, these methods have shown a significant improvement in localization accuracy. As summarized in Table~\ref{table:1} and subdivided into categories in Table \ref{tab:backbone_comparison}, this section delves into an in-depth analysis of several influential deep learning techniques in the field of cross-view geo-localization.

\subsection{Siamese Networks and Pair-based Networks}
A Siamese network~\cite{koch2015siamese} consists of two identical subnetworks that share the same parameters and weights. These subnetworks take two different inputs and process them to extract feature representations. Let \( f(x) \) represent the output of the subnetwork when input \( x \) is passed through it. For two input images \( x_1 \) and \( x_2 \), the Siamese network outputs \( f(x_1) \) and \( f(x_2) \).

The similarity between the two inputs can be measured using a distance metric. One common choice is the Euclidean distance:

\begin{equation}
D(x_1, x_2) = \| f(x_1) - f(x_2) \|_2
\end{equation}

where \( \| \cdot \|_2 \) denotes the Euclidean norm.

The loss function used to train Siamese networks often includes a margin-based contrastive loss, which encourages the network to produce small distances for similar pairs and large distances for dissimilar pairs. The contrastive loss function \( L \) can be defined as:

\begin{equation}
L(y, D) = (1 - y) \frac{1}{2} D^2 + y \frac{1}{2} \{ \max(0, m - D) \}^2
\end{equation}

where \( y \) is a binary label indicating whether the pair is similar (0) or dissimilar (1), and \( m \) is a margin parameter. The first term penalizes similar pairs with large distances, and the second term penalizes dissimilar pairs with distances less than the margin \( m \).

\begin{figure}
\includegraphics[width=\linewidth]{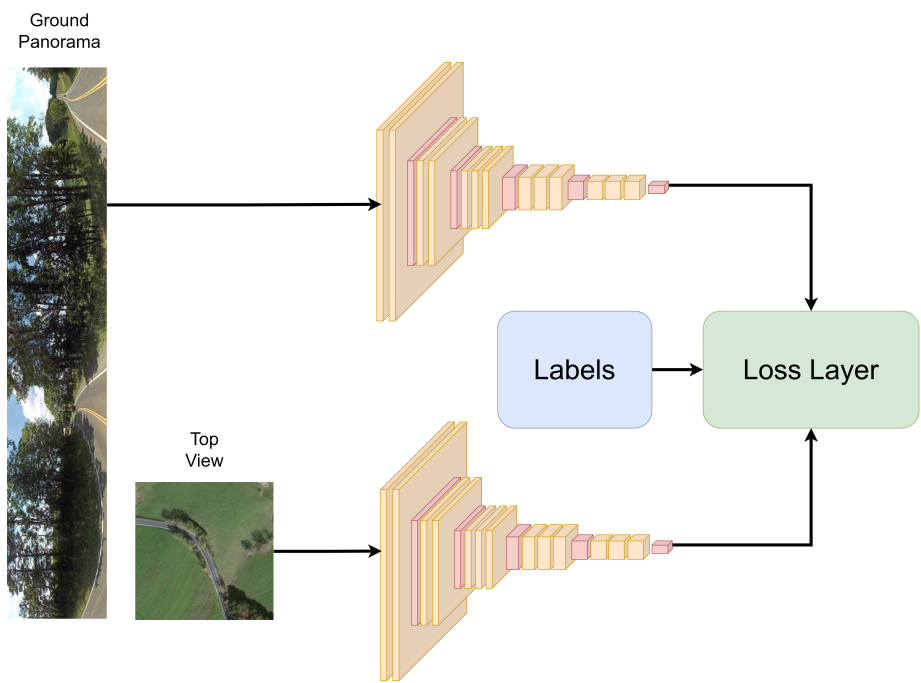}
\caption{An illustrative example of training a pair of Neural Networks, similar to what was proposed in \citet{lin2015learning}. }\label{fig:siamese_intro}
\end{figure}

Siamese networks, along with pair-based network structures, have gained significant traction in the domain of cross-view geo-localization tasks \citep{liu2019lending,tian2017cross,shi2019spatial, hu2018cvm,rodrigues2021these,zhu2021vigor,vyas2022gama}. Crafted specifically to compute similarity or distance metrics between pairs of images or features, they have proven adept for matching images captured from distinct viewpoints, such as ground-level and aerial views \citep{koch2015siamese}. Refer to Figure \ref{fig:siamese_intro}, which provides an illustrative example of training a pair of Neural Networks akin to the model proposed in \citet{lin2015learning}.

Siamese networks are specialized types of neural network architectures that focus on learning similarity or distance metrics between pairs of input instances. Their primary goal is to analyze and discern the relational attributes between different instances of data.

In the sphere of deep learning networks, triplet loss is often employed as a popular loss function aimed at optimizing the network's performance. This loss function is designed to learn a discriminative feature representation by examining the relationships between three distinct input instances: an anchor, a positive, and a negative instance. The ultimate goal of triplet loss is to minimize the distance between the anchor $A$ and the positive instances $P$ in the feature space while maximizing the distance between the anchor and negative instances $N$, all within a predefined margin.
\begin{equation}
L_\text{triplet} = \max \left(0, D(A, P) - D(A, N) + \text{margin}\right),
\label{eq:triplet-loss}
\end{equation}

\noindent where  $D(A, P)$ is the distance between the anchor $A$ and the positive instance $P$, and $D(A, N)$ is the distance between the anchor $A$ and the negative instance $N$. The margin is a predefined value to determine the difference that should exist between the two distances for the loss to be minimized. Note that if the difference between $D(A, P)$ and $D(A, N)$ is larger than the margin, the loss will be zero, indicating that the network has correctly learned the discriminative features.

\citet{vo2016localizing} presented deep learning techniques and dataset specifically engineered to tackle the challenging task of matching ground-level images to overhead imagery. This approach, therefore, facilitated large-scale image geo-localization by making use of readily available overhead or satellite imagery to estimate the location of ground-level photographs.

\subsection{Temporal Independence-Based Methods}
\citet{rodrigues2021these} addressed the challenge of temporal variations in scenes. The primary contribution of the paper lay in the development of a novel data augmentation pipeline. This pipeline enabled the system to handle and match objects not seen during the training phase, providing a significant advantage for images captured at different times or in areas that were occluded or shadowed. The idea of temporal independence was later adopted by \citet{ghanem2023leveraging} in an ensemble model and in \citet{mithun2023cross} for outdoor augmented reality. Furthermore, \citet{rodrigues2021these} introduced a multi-scale attention module to assist in the image-matching task. Unlike previous models that operated solely at a single scale, this attention module functioned at multiple scales, thereby enhancing the overall performance in matching images across different views. The paper highlighted the potential of both data augmentation techniques and multi-scale attention modules for advancing the state of the art in cross-view image geo-localization, especially when temporal variations were a concern.

\begin{longtable}{|p{4.5cm}|p{5.5cm}|l|}
\caption{Ideas introduced by major papers in cross-view Geo-localization.} \label{table:1} \\
\hline
\textbf{Reference \& Paper} & \textbf{Highlights} & \textbf{Publication} \\ \hline
\endfirsthead

\multicolumn{3}{c}%
{{\bfseries \tablename\ \thetable{} -- continued from previous page}} \\
\hline
\textbf{Reference \& Paper} & \textbf{Highlights} & \textbf{Publication} \\ \hline
\endhead

\hline \multicolumn{3}{|r|}{{Continued on next page}} \\ \hline
\endfoot

\hline
\endlastfoot

\citet{vo2016localizing} - ``Localizing and orienting street views using overhead imagery" & Distance-based logistic loss with orientation regression. & CVPR 2016 \\ \hline
\citet{zhai2017predicting} - ``Predicting ground-level scene layout from aerial imagery" & Semantic extraction using adaptive CNN without manual labeling. & CVPR 2017 \\ \hline
\citet{hu2018cvm} - ``CVM-Net: Cross-view matching network for image-based ground-to-aerial geo-localization" & Siamese net with NetVLAD and weighted ranking loss. & CVPR 2018 \\ \hline
\citet{Liu_2019_CVPR} - "Lending Orientation to Neural Networks for Cross-View Geo-Localization" & Pixel orientation in Siamese net for matching to satellite images. & CVPR 2019 \\ \hline
\citet{regmi2019bridging} - ``Bridging the domain gap for ground-to-aerial image matching" & View synthesis via cGANs using segmentation maps. & CVPR 2019 \\ \hline
\citet{shi2019spatial} - ``Spatial-aware feature aggregation for image based cross-view geo-localization" & Spatial correspondence via polar transform and attention. & CVPR 2019 \\ \hline
\citet{shi2020optimal} - ``Optimal feature transport for cross-view image geo-localization" & Cross-view alignment using Sinkhorn operation. & CVPR 2020 \\ \hline
\citet{shi2020looking} - ``Where am i looking at? joint location and orientation estimation by cross-view matching" & Aerial-to-ground alignment with DSM. & CVPR 2020 \\ \hline
\citet{toker2021coming} - ``Coming down to earth: Satellite-to-street view synthesis for geo-localization" & Street synthesis from satellites for cross-view matching. & CVPR 2021 \\ \hline
\citet{yang2021cross} - ``Cross-view geo-localization with evolving transformer" & L2LTR model with self-cross attention in transformers. & CVPR 2021 \\ \hline
\citet{wang2021each} - ``Each part matters: Local patterns facilitate cross-view geo-localization" & LPN model with square-ring partition strategy. & CVPR 2021 \\ \hline
\citet{lin2022joint} - ``Joint Representation Learning and Keypoint Detection for Cross-view Geo-localization" & RK-Net with keypoint focus using USAM. & CVPR 2022 \\ \hline
\citet{zhu2022transgeo} - ``Transgeo: Transformer is all you need for cross-view image geo-localization" & Pure transformer with non-uniform cropping. & CVPR 2022 \\ \hline
\citet{wang2022transformer} - ``Transformer-Guided Convolutional Neural Network for Cross-View Geolocalization" & TransGCNN with multi-scale window transformer. & CVPR 2022 \\ \hline
\citet{wang2022learning} - ``Learning Cross-view Geo-localization Embeddings via Dynamic Weighted Decorrelation Regularization" & DWDR with symmetric sampling for cross-view balance. & CVPR 2022 \\ \hline
\citet{zhao2022mutual} - ``Mutual Generative Transformer Learning for Cross-view Geo-localization" & MGTL with cascaded attention and G2S/S2G sub-modules. & CVPR 2022 \\ \hline
\citet{rodrigues2023semgeo} - ``SemGeo: Semantic Keywords for Cross-View Image Geo-Localization" & Used a CNN to generate keywords from semantic segmented ground view image and these keywords are used as tokens for the transformer & CVPR 2023 \\ \hline
\citet{mi2024congeo} - ``ConGeo: Robust Cross-view Geo-localization across Ground View Variations" & improved a model’s invariance to orientation and its resilience to FoV variations & CVPR 2024 \\ \hline
\end{longtable}

\subsection{Incorporating Orientation Information}
Ground-to-aerial image alignment had gained popularity as an approach for solving image-based geo-localization problems, largely due to the widespread availability and extensive coverage provided by satellite imagery.

Given an image captured by a ground-level camera, two essential questions naturally arise: where was the camera located, and in which direction was it facing?

Understanding the orientation relationship between ground and aerial images could considerably mitigate the ambiguity in matching these two types of views. This was especially true when the ground-level images offered a limited Field of View (FoV), as opposed to a full panoramic view.

Nevertheless, cross-view alignment remained a challenging task, primarily because of the drastic differences in viewpoints between ground-level and aerial images. One way to bridge this gap was by transforming the aerial image into a polar representation. Polar transform was first seen in \citet{shi2019spatial} and later adopted in different methods \citep{shi2020looking,wang2022learning,toker2021coming}. In a work by \citet{li2023multi}, a inverse polar transform was applied.

Let the polar coordinates be \( [x_{si}^{s}, y_{si}^s] \). The coordinates are calculated through a combination of scaling and shifting operations on the \( x \)-coordinate and \( y \)-coordinate of the original satellite image with coordinates \([x_{si}, y_{si}]\). Here, \( x_{si} \) and \( y_{si} \) represent the original \( x \)- and \( y \)-coordinates of the satellite image. The scaling factor involves the ratio of the polar-transformed image's \( y \)-coordinate (\( y_{psi}^{ps} \)) to its height (\( H_{ps} \)). The shifting component is determined by the half-width of the satellite image (\( W_s/2 \)). Additionally, a sine function is used to introduce a rotational effect, influenced by the \( x \)-coordinate in the polar-transformed image (\( x_{psi}^{ps} \)).

\begin{align}
\begin{bmatrix} x_{si}^{s}\\ y_{si}^{s}\end{bmatrix} = 
\begin{bmatrix} \frac{W_s}{2} \\ \frac{H_s}{2} \end{bmatrix}+ \begin{bmatrix} 
\frac{W_s}{2} \cdot \frac{y_{psi}^{ps}}{H_{ps}} \cdot \sin\left(\frac{2\pi}{W_s} \cdot x_{psi}^{ps}\right)    
\\
- \frac{H_s}{2} \cdot \frac{y_{psi}^{ps}}{H_{ps}} \cdot \cos\left(\frac{2\pi}{W_s} \cdot x_{psi}^{ps}\right)     
\end{bmatrix}
\label{eq:aerial-to-polar}
\end{align}

These equations execute the requisite transformations to convert pixel coordinates from the satellite image to a polar-transformed image. In this new coordinate system, circular lines in the satellite image are transformed into horizontal lines, while radial lines become vertical. Notably, the north-line, which originates from the satellite image's center, aligns with the vertical line at \( W_{ps}/2 \) in the transformed image. It is crucial to note, however, that while this transformation retains the general layout of objects in the scene, it may not fully eliminate the domain gap between the two perspectives.

Following this transformation, models such as dynamic similarity matching \citep{shi2020looking} Figure \ref{fig:dsm_model} can be employed to compute the correlation between these transformed and ground-level views.

The Spatial-aware Feature Aggregation (SAFA) \citep{shi2019spatial} technique for the task of ground to aerial cross-view image-based geo-localization is another notable method which makes use of polar transformation. This methodology specifically addressed the challenges of image distortions in aerial views and the importance of spatial configuration in feature representation which when integrated with other methods \citep{wang2021each,zhang2023cross,lin2022joint}.  The strategy leveraged an initial polar transform to align the domains of ground and aerial view images, followed by a SAFA module that re-weighted features, embedding the spatial configuration and providing a robust and discriminative feature representation. A key component of the SAFA module is the Spatial-aware Position Embedding Module (SPE), which employs a self-attention mechanism under a metric learning objective. The SPE encodes the relative positions among object features and selects distinct features, mitigating the impact of distortions caused by the polar transform. The robustness of the feature representation is further enhanced by a strategy of Multiple Position-embedded Feature Aggregation, using multiple SPEs to generate different embedding maps that focus on different spatial layouts. The paper uses a weighted soft-margin triplet loss function for training, aiming to bring matching pairs closer while pushing unmatching pairs apart in the feature space. 

The advent of color-coded orientation maps marked a significant milestone in the realms of image processing, computer vision, and remote sensing \citep{liu2019lending}. An orientation map can be understood as a visualization tool where each point (or pixel) on the map represents the orientation, see figure 3 (e.g., azimuth and altitude or azimuth and range for ground-level panoramas and satellite images, respectively) at the corresponding location in the space. This is represented using hue and saturation channels in a color map, facilitating the intuitive and rapid interpretation of orientation information which otherwise tends to be elusive when gleaned from raw data sets. The incorporation of these orientation maps into neural network models furnished several advantages. Most notably, they allowed the network to more effectively harness the directional information embedded within images. This, in turn, contributed to generating predictions that were both more robust and accurate. Moreover, the use of orientation maps also offered computational efficiency, obviating the necessity for complex number computations as was the case with complex-valued neural networks.

\begin{figure}
\begin{center}
\includegraphics[width=\linewidth]{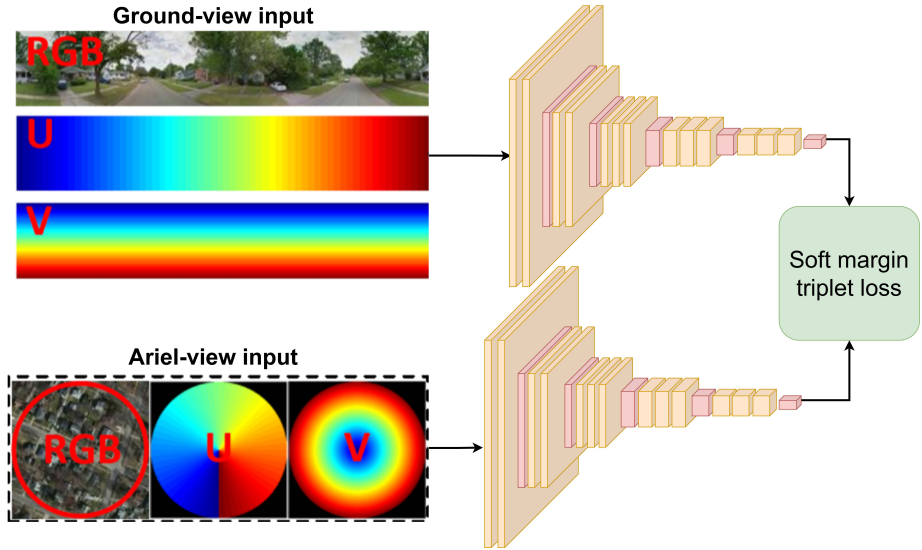}
\caption{Encoding orientation using color maps u,v for altitude and azimuth as proposed by \citet{Liu_2019_CVPR}.}\label{fig:orientation_estimation}
\end{center}
\end{figure}

\begin{figure*}
\begin{center}
\includegraphics[width=\linewidth]{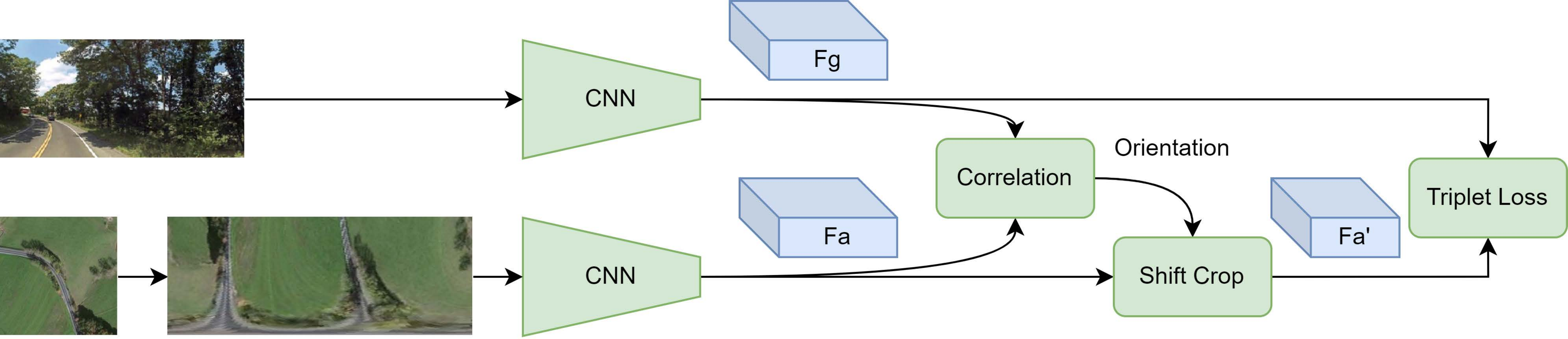}
\caption{The flowchart of geo-localization method proposed by \citet{DBLP:journals/corr/abs-2005-03860}. Initially, an aerial image undergoes a polar transformation. Following this, a dual-stream CNN is employed to derive features from both ground-level and polar-transformed aerial images. The features extracted serve to compute the correlation between the two viewpoints, facilitating the estimation of the ground image orientation in relation to its aerial counterpart. Subsequent to this process, the features originating from the aerial perspective are adjusted and trimmed to align with the potential area corresponding with ground view features. The similarities identified amongst the adjusted features are then leveraged for the purpose of location retrieval.}\label{fig:dsm_model}
\end{center}
\end{figure*}

In summary orientation maps and their color-coded counterparts have emerged as pivotal tools in seamlessly incorporating and visualizing orientation data across diverse image-driven applications. Their nuanced methodologies have simplified the intricate task of interpreting orientation data, invariably enhancing the efficacy of image-based computational tasks. A notable application of this approach involves encoding orientation details as two supplementary color channels, often termed the U-V channels, juxtaposed with the conventional 3-channels RGB input. This enriched format effortlessly integrates into existing models, such as Siamese networks, amplifying their performance. For a visual elucidation of this encoding technique, which employs color maps u and v to represent altitude and azimuth respectively, one can refer to Figure \ref{fig:orientation_estimation}, as proposed by \citet{Liu_2019_CVPR}.

Another significant contribution was presented in \citet{cai2019ground} is the introduction of a Fully Convolutional Attention Module (FCAM). This lightweight attention mechanism is seamlessly integrated into a foundational residual network. Siamese networks are characterized by their architecture, which comprises two or more identical subnetworks. This specialized architecture, when combined with the FCAM, outperforms its counterparts that lack this attention module, thereby augmenting the model's overall effectiveness significantly.

In summary, these strategic innovations \citep{liu2019lending,cai2019ground,shi2019spatial,wang2021each,zhang2023cross,lin2022joint} synergize to forge a highly proficient system for cross-view image geo-localization. This system is optimized to match ground-level images with their aerial counterparts, thereby surmounting a multitude of challenges such as extreme viewpoint differences, varying lighting conditions, and uncertainties in orientation that typically plague the matching process between ground and aerial images. As a result, this system showed marked improvements over existing state-of-the-art methodologies.

\subsection{Capsule Networks}
Capsule Networks (CapsNets) \citep{sun2019geocapsnet,zhu2021geographic} have demonstrated potential in addressing the complexities associated with aerial-to-ground view image geo-localization. Their capability to encode not only the features but also their spatial relationships allows for effective handling of viewpoint changes, scale differences, and occlusions. The concept of using Capsule Networks for geo-localization was first proposed in  \citet{sun2019geocapsnet}, which also introduced the Soft-TriHard Loss. This is a modified version of the conventional Triplet Loss, designed to improve the training of deep embedding models by promoting the learning of more discriminative feature representations.

In the GeoCapsNet architecture, which is highlighted in the aforementioned literature, two layers of capsules are employed: PrimaryCaps and GeoCaps. The PrimaryCaps layer takes basic features detected by a feature extractor (such as ResNetX in this context) and combines them to form a set of higher-level features. These primary capsules are then processed through the GeoCaps layer, which encodes the relative spatial relationships between these higher-level features, thus furnishing a robust image representation suitable for cross-view image matching tasks.

\begin{equation}
L_{sth} = \frac{1}{M} \sum_{a \in \text{batch}} \ln \left( 1 + e^{\alpha (d_{a,p} - \min_{n \in B} d_{a,n})} \right)
\end{equation}
where \( L_{sth} \) is the Soft-TriHard Loss, \( M \) is the number of samples in the mini-batch, \( d_{a,p} \) is the distance between the capsule feature of the anchor \( a \) and its positive sample \( p \), \( d_{a,n} \) is the distance between the capsule feature of the anchor \( a \) and its negative sample \( n \), \( \alpha \) is a scaling factor, and \( B \) is the set of negative samples within the batch. The Soft-TriHard Loss is designed to accelerate the model's convergence rate by selecting the hardest negative samples in a mini-batch and utilizing a weighted soft-margin ranking loss.

\subsection{Part-based Representation Learning}
Part-based representation learning is an approach in geo-localization that seeks to enhance the performance of localization tasks by concentrating on distinct parts or regions of an image. This methodology subdivides images into smaller segments or patches, empowering the model to learn more granular, discriminative, and resilient feature representations for each segment. This strategy effectively addresses challenges such as viewpoint variations, occlusions, and illumination changes, which are frequently encountered in cross-view geo-localization tasks. This strategy involves, dividing images into smaller segments, patches, or regions to allow the model to focus on local patterns and structures, thereby facilitating the matching of different views \citep{wang2021each}. Specifically, square ring partitions are applied to satellite images and row partitions to ground panoramas. In the case of satellite images, they are divided into multiple concentric square-shaped rings, capturing local patterns at varying distances from the image center. This partitioning technique ensures that the features extracted are both distinctive and robust, considering the spatial context of local patterns within the image. As for the ground panorama images, they are divided into multiple rows or horizontal strips, with each strip capturing a specific range of viewing angles from the ground-level perspective. This row-based partitioning simplifies the representation of ground panoramas and enables the extraction of local features that are highly discriminative and useful for cross-view matching tasks.

\begin{figure*}
\begin{center}
\includegraphics[width=\linewidth]{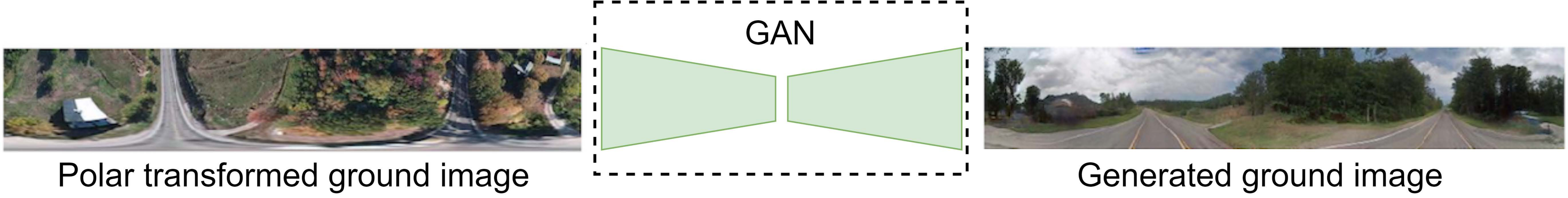}
\caption{The schematic showcases the GAN-based synthesis of ground view images, beginning with an initial polar transformed image (on the left) and culminating in a GAN-generated ground view (on the right) \citep{toker2021coming}}\label{fig:GAN_cross_view}
\end{center}
\end{figure*}

\subsection{Generative Adversarial Networks based methods}

Generative Adversarial Networks (GANs) have solidified their standing as a formidable tool in the realm of cross-view geo-localization \citep{toker2021coming,wu2022cross,regmi2018cross,regmi2019cross,regmi2019bridging, tang2019multi}. Their prowess is especially evident when crafting models tailored for niche applications. A groundbreaking study \citet{toker2021coming} unveiled a versatile architecture adept at juggling the dual challenges of image synthesis and retrieval. Integral to this architecture is the GAN's role in converting satellite data into realistic ground-level imagery, thereby bridging the visual chasm that typically exists between aerial and terrestrial perspectives. Figure \ref{fig:GAN_cross_view} offers a illustration of this process, demonstrating how a polar transformed ground image undergoes a metamorphosis to yield a detailed ground view image via the GAN. The underlying mechanism of the GAN hinges on its capability to identify and harness features characteristic to both views, bolstering the accuracy of image retrieval in the process. In tandem with this, the network's intrinsic retrieval function steers the GAN, ensuring the synthesized ground-level visuals resonate with the essence of the corresponding satellite imagery.

A key innovation of this approach is the use of polar transformed satellite images as an initial point for the GAN. This choice simplifies the image generation task as the spatial layout in the polar transformed images closely resembles that of street views, thereby reducing the complexity of the required transformations.

\subsection{Transformer Methods and Attention-Based Models}
Transformers \citep{vaswani2017attention} are a type of deep learning model that has gained significant attention in recent years due to their effectiveness in various fields, particularly in natural language processing. Their ability to handle long-term dependencies and capture global context in data has made them especially useful. Transformers employ a self-attention mechanism, allowing them to weigh and combine all parts of the input data. This eliminates the need for sequential processing, which is particularly beneficial in tasks like cross-view geo-localization. In recent years, a large number of models have utilized attention in one way or another \citep{zhu2023simple,wang2023satellite,wang2023cross,xu2023aenet,li2023multi,zhang2021ssa,yang2021cross,zhu2022transgeo,zhang2023cross,lu2022content,fervers2023uncertainty,zhao2022co,cao2019learning,wang2023dehi,tian2022smdt}.

The primary contribution of the work by \citet{yang2021cross} lies in the development of the EgoTR model. It is the first of its kind to employ a Transformer for cross-view geo-localization tasks. By being the first to apply a Transformer to this specific problem \citep{yang2021cross}, EgoTR leverages the globally context-aware nature of the architecture to mitigate visual ambiguities—a common challenge in cross-view geo-localization. Additionally, the model incorporates geometric concepts through positional encoding to reduce ambiguities based on geometric misalignment. Uniquely, EgoTR learns position embeddings without imposing strong assumptions about position knowledge, enhancing its applicability in practical scenarios. A key innovation in EgoTR is the proposed self-cross attention mechanism, designed to facilitate effective information flow across Transformer blocks and promote the evolution of representations. This enhancement boosts EgoTR's representational and generalization abilities without incurring additional computational costs. Extensive experiments demonstrate that the EgoTR model brings consistent and substantial improvements across various cross-view matching tasks, achieving state-of-the-art performance.

Besides, \citet{zhu2022transgeo} introduced TransGeo—a completely transformer-based approach to cross-view image geo-localization. Unlike EgoTR, which contains elements of CNNs, TransGeo neither relies on polar transforms nor data augmentation, marking a significant departure from traditional methods. The paper also introduces a new attention-guided non-uniform cropping strategy. This technique effectively reduces the number of non-informative patches in reference aerial images, thereby lowering computational overhead while maintaining performance. Further optimization repurposes the saved computation to enhance the resolution of key informative areas in images. TransGeo exhibits lower computational cost, GPU memory usage, and inference time than its CNN-based counterparts, setting new standards in cross-view geo-localization.

\begin{figure}
\begin{center}
\includegraphics[scale=0.05]{fig6.pdf}
\caption{Workflow from TransGeo method \citep{zhu2022transgeo}.}\label{fig:tgoe}.
\end{center}
\end{figure}

For an in-depth understanding of the TransGeo methodology, Figure \ref{fig:tgoe} provides a detailed schematic of the process, and Fig. \ref{fig12.png} shows an example of the attention map of the TransGeo model. TransGeo incorporates a sophisticated non-uniform cropping algorithm, underpinned by attention-based mechanisms. This innovative strategy selectively excises the less informative segments within aerial imagery, thereby optimizing the computational expenditure. The savings in computational resources are judiciously redirected towards enhancing the resolution of salient regions, thereby amplifying the overall performance of the model. Such a focused approach does not solely confer efficiency gains; it substantially upgrades the fidelity of feature representation, which, in turn, significantly refines the accuracy of geo-localization outcomes.

\begin{figure}
\begin{center}
\includegraphics[scale=0.7]{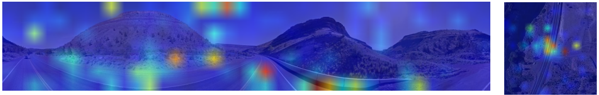}
\caption{Attention in TransGeo model \citep{zhu2022transgeo}.}\label{fig:tgoe_attn}.
\end{center}
\end{figure}

In the domain of geo-spatial analysis, it is well-acknowledged that not all areas within an image hold equal value for localization purposes. Regions endowed with unique landmarks or distinct topographical features serve as pivotal navigation markers. On the other hand, monotonous landscapes or areas with repetitive architectural patterns offer limited locational insights. Leveraging this heterogeneity, non-uniform cropping utilizes an astute attention-guided mechanism to discern and prioritize patches that are densely packed with information, thereby streamlining the model’s focus and processing capabilities.

\citet{wang2022transformer} developed a hybrid network that integrates a Transformer-style head network as a spatial attention module. This module selects significant CNN features for the final image descriptor, thereby while ensuring efficiency and lightness, the embedded features possess discriminative capability. A unique multi-scale window Transformer is proposed by the authors. This innovative approach conducts self-attention on the input map’s multi-scale windows independently and dynamically combines this multi-scale information, which serves to enhance the global representation effectively. Extensive experiments demonstrate an advantageous balance between accuracy and efficiency, as the model outperforms the second-best model, EgoTR, with fewer parameters and a higher frame rate.

The Mutual Guided Transformer Learning (MGTL) approach \citep{zhao2022mutual} represents a significant advancement in Cross-view Geo-localization (CVGL). Unique to this approach is the focus on mutual interaction between ground-level and aerial-level patterns, going beyond the self-attentive reasoning found in existing transformer-based CVGL models. This is the first time such interaction has been attempted in the CVGL field, to the best of our knowledge.

Another innovation in MGTL is a cascaded attention-guided masking technique, which leverages co-visual patterns for improved performance. Unlike traditional methods that treat aerial and ground views equivalently, MGTL employs an attention-guided exploration algorithm, enhancing the model's performance.

The effectiveness of these novel developments is demonstrated by the state-of-the-art localization accuracy achieved on widely-used benchmarks such as CVUSA and CVACT, outperforming existing deep learning models.

The Simple Attention-based Image Geo-localization backbone (SAIG) \citep{zhu2023simple} is designed to serve as a simple yet effective model backbone for cross-view geo-localization. It incorporates an attention mechanism skilled in capturing long-range information, crucial for feature matching. The model's design enables it to bridge the domain gap and correspond to global features without imposing strong assumptions. With only 18.2 million parameters in its smaller model compared to TransGeo's 44.9 million parameters \citet{zhu2022transgeo}, SAIG offers a compact design.

Additionally, \citet{zhu2023simple} propose a new spatial-mixed feature aggregation module and two losses to enhance SAIG's performance in one-to-many scenarios. The model's effectiveness has been demonstrated through comprehensive experiments and ablation studies in various cross-view geo-localization and image retrieval settings.


\begin{longtable}{|p{2cm}|p{3cm}|p{6cm}|}
\caption{Comparison of different backbones used in various feature types} \label{tab:backbone_comparison} \\
\hline
\textbf{Feature Type} & \textbf{Backbone} & \textbf{Relevant articles} \\ \hline
\endfirsthead

\multicolumn{3}{c}%
{{\bfseries \tablename\ \thetable{} -- continued from previous page}} \\
\hline
\textbf{Feature Type} & \textbf{Backbone} & \textbf{Used in} \\ \hline
\endhead

\hline \multicolumn{3}{|r|}{{Continued on next page}} \\ \hline
\endfoot

\hline
\endlastfoot
Handcrafted & SIFT & \citet{zemene2018large, zamir2014image} \\
\cline{2-3} 
 & SURF, FREAK, PHOW & \citet{zemene2018large} \\
\cline{2-3} 
 & SIFT + VLAD & \citet{middelberg2014scalable} \\
\hline
Semantic & Faster R-CNN & \citet{tian2017cross} \\
\hline
CNN & VGG + FCN + NetVLAD & \citet{hu2018cvm, zhu2022transgeo, wang2021each} \\
\cline{2-3} 
 & AlexNet & \citet{vo2016localizing, tian2017cross} \\
\cline{2-3} 
 & VGG & \citet{Liu_2019_CVPR, tian2017cross, samano2020you, shi2020optimal} \\
\cline{2-3} 
 & ResNet & \citet{Liu_2019_CVPR, tian2017cross, rajasegaran2019deepcaps} \\
\cline{2-3} 
 & DenseNet & \citet{Liu_2019_CVPR, tian2017cross} \\
\cline{2-3} 
 & U-Net & \citet{Liu_2019_CVPR} \\
\cline{2-3} 
 & Xception & \citet{tian2017cross} \\
\cline{2-3} 
 & VGG + FCN & \citet{zhu2021revisiting} \\
\hline
Attentive & Siam-FCANet (ResNet + FCAM + FCN) & \citet{cai2019ground} \\
\cline{2-3} 
 & Siam-VFCNet (ResNet + FCAM + NetVLAD) & \citet{cai2019ground} \\
\cline{2-3} 
 & VGG + SAFA + SPE & \citet{shi2019spatial, zhu2023simple, lin2022joint} \\
\cline{2-3} 
 & VGG + Geo Attention + Geo-temporal Attention & \citet{regmi2019cross} \\
\cline{2-3} 
 & ResNet + Self Cross Attention & \citet{rodrigues2021these} \\
\cline{2-3} 
 & SAFA & \citet{zhu2021vigor} \\
\cline{2-3} 
 & ResNet + SAFA & \citet{toker2021coming} \\
\cline{2-3} 
 & ResNet + Self Cross Attention & \citet{yang2021cross} \\
\cline{2-3} 
 & VGG + MSAE & \citet{zhu2022transgeo} \\
\hline
Synthesized & X-Fork & \citet{regmi2019cross} \\
\hline

\end{longtable}

\section{Datasets for Cross-view Geo-localization}

\subsection{Ground View Datasets}
Ground view datasets have played a pivotal role in the advancement and assessment of algorithms in the field of geo-localization. Below are some of the noteworthy ground view datasets used in this domain.

\textbf{Pittsburgh 250k Dataset} \citep{torii2013visual}: The Pittsburgh 250k dataset is a large-scale, ground-level, image-based geo-localization dataset originating from the city of Pittsburgh, Pennsylvania. It comprises approximately 250,000 images, each associated with its respective GPS coordinates, giving rise to the name Pittsburgh 250k. Created by researchers at Carnegie Mellon University, this dataset aims to facilitate the development and evaluation of geo-localization algorithms.

The dataset, sourced from Google Street View, encompasses a wide range of urban settings, including residential, commercial, and industrial areas. Due to its large volume and diverse selection of ground-level images, the Pittsburgh 250k dataset serves as an invaluable resource for scholars and practitioners in the field of geo-localization. Its comprehensiveness makes it well-suited for both the training and testing of machine learning algorithms tailored for image-based localization tasks.

\textbf{Google Street View Dataset} \citep{zamir2014visual}: This dataset contains 62,058 high-quality images covering downtown areas and adjacent regions in cities such as Pittsburgh, PA; Orlando, FL; and parts of Manhattan, NY. Each image in the dataset comes with accurate GPS coordinates and compass directions, making it a highly valuable asset for researchers in the field of geo-localization.

The dataset subdivides every 360° Street View placemark into four side views and one upward view. Additionally, each placemark includes an image with overlaid markers indicating addresses, street names, and other pertinent information. A figure in the dataset displays a selection of sample Street View images from eight different placemarks on the left side, while the right side showcases 16 user-uploaded images that were used as query images in the associated paper.

The Google Street View dataset serves as an excellent platform for the development and evaluation of image-based geo-localization algorithms. Its diverse set of images, along with precise GPS coordinates and compass directions, enables robust training and testing across various urban environments. The inclusion of user-uploaded query images further enriches the dataset, allowing for more rigorous evaluation of an algorithm's robustness and accuracy in real-world conditions.

\begin{table}[htbp]
\centering
\caption{Comparison of CVUSA \citep{workman2015localize}, CVACT \citep{liu2019lending}, and VIGOR \citep{zhu2021vigor} datasets.} \label{tab:dataset_comparison}
\begin{tabularx}{\linewidth}{|X|X|X|X|X|X|}
\hline
 & Ground-view image res. & GPS-tag & Satellite resolution & \#training & \#testing \\
\hline
CVACT & 1664x832 & Yes & 1200x1200 & 35,532 & 92,802 \\
\hline
CVUSA & 1232x224 & No & 750x750 & 35,532 & 8,884 \\
\hline
VIGOR & 2048x1024 & Yes & ~0.114 m & 105,214 & - \\
\hline
\end{tabularx}
\end{table}

\subsection{CVUSA}
The CVUSA (Cross-View USA) dataset  \citep{workman2015localize} serves as a large-scale resource for researchers working on cross-view geo-localization. The dataset comprises 10,000 street-level images sourced from 20 cities across the United States, including major urban centers like New York, San Francisco, and Los Angeles. These images, collected from Google Street View, depict a broad spectrum of urban scenes, such as residential neighborhoods, commercial districts, and landmarks.

For each image, the CVUSA dataset provides ground-truth camera poses and GPS coordinates. It also includes a set of query images designed to test the robustness of cross-view geo-localization and place recognition algorithms. These queries span various challenges, including changes in weather, lighting conditions, occlusions, and differing viewpoints. Due to the large variations in scene appearance and lighting conditions, the dataset poses a considerable challenge for researchers.

CVUSA has rapidly become a standard benchmark for evaluating algorithms in the domain of cross-view geo-localization and place recognition. Several research publications have already employed this dataset to advance the state-of-the-art methodologies in this field.

\begin{figure*}
\begin{center}
\includegraphics[scale=0.37]{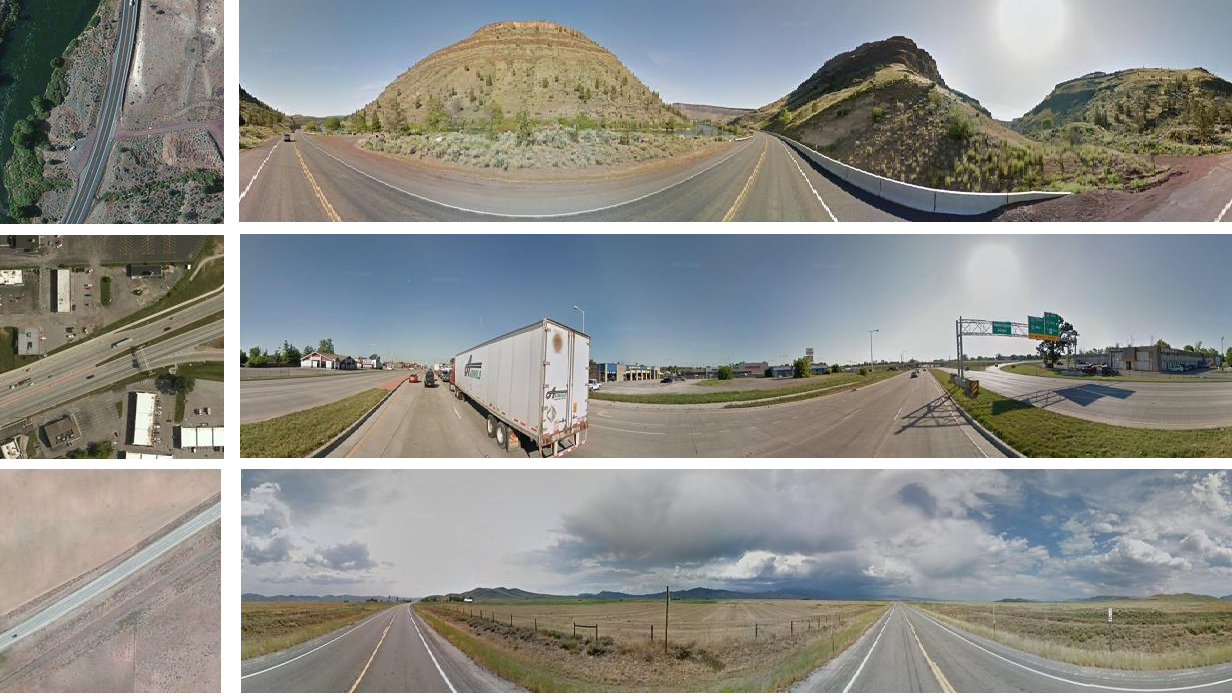}
\caption{Example images from the CVUSA dataset \citep{workman2015localize}. The image on the left is a panoramic street-level view, while the one on the right shows the corresponding aerial perspective.}\label{fig:CVUSA}
\end{center}
\end{figure*}

\subsection{CVACT}
The CVACT dataset \citep{liu2019lending} is another large-scale resource tailored for cross-view geo-localization research, specifically for estimating a ground-level image's GPS coordinates based on its appearance and corresponding aerial view. The dataset features ground-level panoramas sourced from Google Street View and high-resolution satellite images from Google Maps, all tagged with accurate GPS coordinates. It densely covers a 300-square-mile area around the city of Canberra at zoom level 2.

The dataset includes 128,334 cross-view image pairs, partitioned into 35,532 training pairs and 92,802 testing pairs. The ground-level panoramas boast a 360-degree field of view and a resolution of 1664 $\times$ 832 pixels, while the satellite images offer a 1200 $\times$ 1200-pixel resolution and a ground resolution of 0.12 meters per pixel. CVACT is open to the public and can be accessed via the project's website.

The CVACT dataset stands as a crucial asset for scholars and practitioners working on a variety of applications, such as autonomous driving, urban planning, and surveillance. Its challenging nature provides an effective testbed for different algorithms and can significantly contribute to the advancement of cross-view geo-localization research.

\begin{figure*}
\begin{center}
\includegraphics[scale=0.17]{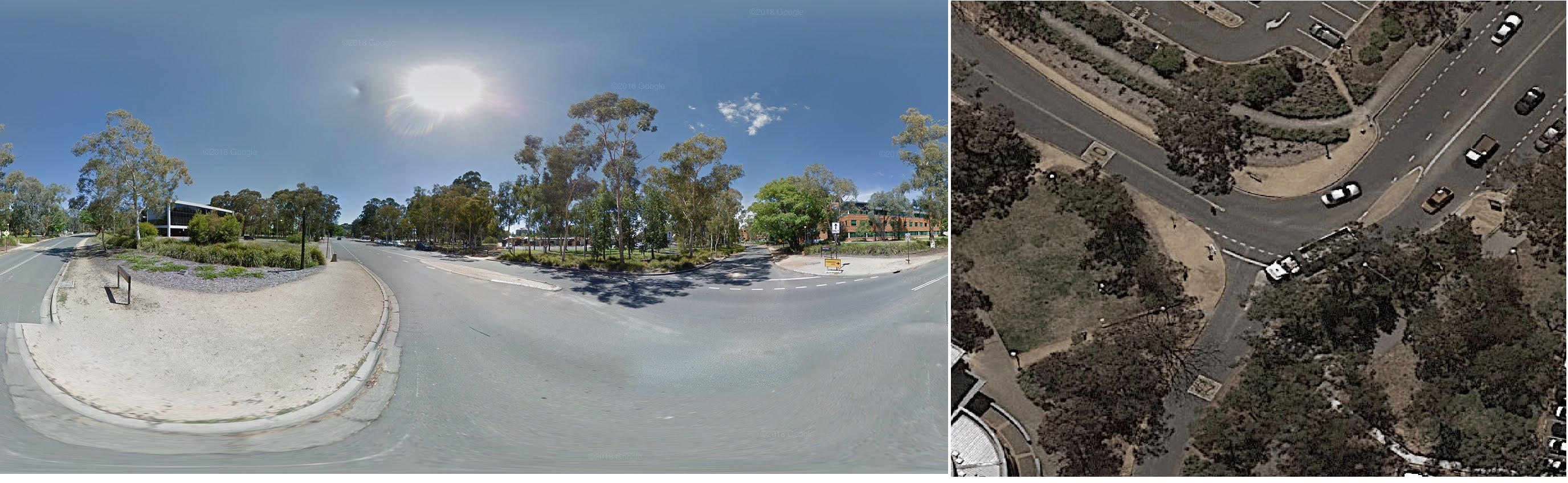}
\caption{Sample images from CVACT dataset \citep{liu2019lending}. Example image to the left is a panoramic image and to the right is the corresponding top view image.}\label{fig:CVACT_dataset_example}
\end{center}
\end{figure*}

\subsection{VIGOR}
The VIGOR dataset, as referenced in \citet{zhu2021vigor}, is a comprehensive collection curated for cross-view geo-localization research. This dataset encompasses aerial photographs and street-level panoramas from four predominant U.S. cities, namely New York City (Manhattan), San Francisco, Chicago, and Seattle. The aerial photographs within the dataset were acquired via the Google Maps Static API, and the street-view panoramas were secured through the use of the Google Street View Static API.

The dataset is unique in that it maintains a consistent interval—typically around 30 meters—between the GPS locations of panorama samples. Care has been taken to balance the dataset, ensuring that each aerial image is associated with no more than two corresponding panoramas.

In this dataset, aerial-view images come in a 640 $\times$ 640-pixel format, and ground-view panoramas are sized at 2048 $\times$ 1024 pixels. The satellite images have a zoom level of 20, translating to a ground resolution of approximately 0.114 meters. All images are tagged with industrial-grade GPS coordinates to facilitate meter-level evaluation.

To further refine the dataset, the orientation of each panorama is adjusted so that North is centered. In addition, about 4\% of the aerial images have no corresponding panoramas; these are intentionally included as distraction samples to present a more challenging dataset for researchers. Table \ref{tab:dataset_comparison} compairs CVUSA \citep{workman2015localize}, CVACT \citep{liu2019lending}, and VIGOR \citep{zhu2021vigor} datasets.

\begin{figure*}
\begin{center}
\includegraphics[scale=1]{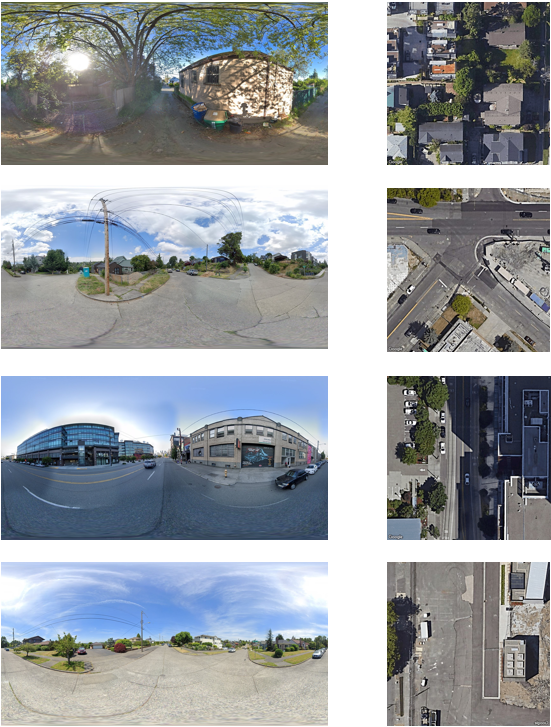}
\caption{Sample images from VIGOR dataset \citep{zhu2021vigor}. Example images from the Seattle collection, showing orientation correspondence between an aerial and a street-view image.}\label{fig:VIGOR}
\end{center}
\end{figure*}

\subsection{VO}
The Vo dataset \citep{vo2016localizing} is expansive, comprising over a million pairs of ground and aerial photos.  Unique from other datasets discussed in this segment, the Vo dataset emphasizes scene localization in the image, assuming the principal object is situated at the satellite image's center. To illustrate, in Figure 26's first column, the building at the satellite image's center corresponds to the primary object in the ground photo. This approach distinguishes it from datasets like CVUSA \citep{workman2015localize} and CVACT \citep{liu2019lending}. In developing this dataset, the authors sourced street-view panoramic shots from GSV \citep{anguelov2010google}, segmenting them into various crops. They then obtained depth estimates for each segment from GSV and secured the associated satellite imagery through the Google Maps API18. The base panoramas were gathered at random from 11 distinct US cities using GSV \citep{anguelov2010google}.

\subsection{UrbanGeo}
The dataset \citet{tian2017cross} emphasized crossview geo-localization within urban settings, gathering data from three American cities: Pittsburgh, Orlando, and a portion of Manhattan. A total of 8,851 GPS points were amassed from these locales by the author. For every GPS point, four distinct bird's-eye view images were taken, oriented at headings of 0°, 90°, 180°, and 270°. The DualMaps20 tool was utilized to secure the corresponding street view images from GSV \citep{anguelov2010google} based on the provided bird’s-eye view picture. Additionally, the researcher marked the boundaries of each building on both the street view and bird's-eye view images when the structure appeared in both perspectives.

\section{Results}

The preceding sections have provided an in-depth exploration of existing methods and datasets in the field of cross-view geo-localization. This section presents a comparative analysis, where we examine the performance of various techniques across different datasets. Our objective is to offer a comparative analysis that illuminates the strengths and weaknesses of each method, while providing insights into their applicability under various conditions and constraints. The results are based on data reported in respective publications.

Through this data, we aim to deliver a clear perspective on the state of the art in cross-view geo-localization. We highlight the progress made to date and identify future research challenges. Subsequent subsections will present detailed results for each method category.

CVUSA and CVACT are among the most frequently used datasets for evaluating cross-view geo-localization methods, owing to their extensive coverage and high-quality annotations. The CVUSA dataset comprises pairs of ground-level panoramic images and corresponding overhead satellite views, encompassing a large portion of the United States. The diversity in regions and environmental conditions allows for a robust evaluation of geo-localization methods across various scenarios.

Conversely, the CVACT dataset offers aerial-to-camera cross-view image pairs captured at different times of day, across varying seasons and cities. This dataset introduces challenges such as variations in illumination and weather conditions, among others. These attributes render it a more complex dataset for evaluation, facilitating a thorough assessment of a method's adaptability and performance under diverse conditions.

In this paper, we primarily utilize these two datasets for evaluating different geo-localization methods. This selection is motivated by the need to maintain consistency in our comparative analysis and to align our assessments with standard practices in the field. These datasets offer a broad range of scenarios, making them ideal for testing the robustness and generalization capabilities of different methods.

\begin{table}
\centering
\caption{Performance evaluation results on CVUSA and CVACT datasets. }
\label{table:performance}
\begin{tabular}{|l|llll|llll|}
\hline
Reference Paper & \multicolumn{4}{c|}{CVUSA} & \multicolumn{4}{c|}{CVACT} \\
& r@1 & r@5 & r@10 & r@1\% & r@1 & r@5 & r@10 & r@1\% \\ \hline
 \citet{workman2015wide} & - & - & - & 34.30 & - & - & - & - \\
 \citet{vo2016localizing} & - & - & - & 63.70 & - & - & - & - \\
 \citet{zhai2017predicting} & - & - & - & 43.20 & - & - & - & - \\
 \citet{hu2018cvm} & 22.47 & 49.98 & 63.18 & 93.62 & 5.41 & 14.79 & 25.63 & 54.53 \\
 \citet{Liu_2019_CVPR} & 40.79 & 66.82 & 76.36 & 96.12 & 19.9 & 34.82 & 41.23 & 63.79 \\
 \citet{regmi2019bridging} & 48.75 & - & 81.27 & 95.98 & - & - & - & - \\
 \citet{shi2019spatial} & 89.84 & 96.93 & 98.14 & 99.64 & 55.50 & 79.94 & 85.08 & 94.49 \\
 \citet{shi2020optimal} & 61.43 & 84.69 & 90.49 & 99.02 & 34.39 & 58.83 & 66.78 & 95.99 \\
 \citet{shi2020looking} & 91.96 & 97.50 & 98.54 & 99.67 & 35.55 & 60.17 & 67.95 & 86.71 \\
 \citet{toker2021coming} & 92.56 & 97.55 & 98.33 & 99.57 & 61.29 & 85.13 & 89.14 & 98.32 \\
 \citet{yang2021cross} & 94.05 & 98.27 & 98.99 & 99.67 & 60.72 & 85.85 & 89.88 & 96.12 \\
 \citet{wang2021each} & 93.78 & 98.50 & 99.03 & 99.72 & - & - & - & - \\
 \citet{lin2022joint} & 90.16 & - & - & 99.67 & 56.16 & - & - & 95.22 \\
 \citet{lin2022joint} & 91.22 & - & - & 99.67 & 37.71 & - & - & 87.04 \\
 \citet{zhu2022transgeo} & 94.08 & 98.36 & 99.04 & 99.77 & - & - & - & - \\
 \citet{wang2022transformer} & 94.15 & 98.21 & 98.94 & 99.79 & - & - & - & - \\
 \citet{wang2022learning} & 94.33 & 98.54 & 99.09 & 99.80 & - & - & - & - \\
 \citet{zhao2022mutual} & 94.50 & 98.41 & 99.20 & 99.78 & 61.55 & 86.61 & 90.74 & 98.461 \\
 \cite{zhang2023geodtr+} & 95.40 & 98.44 & 99.05 & 99.75 & 67.57 & 89.84 & 92.57 & 98.54 \\ \hline
\end{tabular}
\end{table}

The comprehensive table presented as Table~\ref{table:performance} summarizes the remarkable progress in the field of cross-view geo-localization, specifically showcasing the advancements in models over the years on the CVUSA and CVACT datasets. Early models like those of  work by \citet{workman2015wide} and \citet{vo2016localizing} showed preliminary results, primarily on the CVUSA dataset. However, recent developments such as LPN+DWDR \citep{wang2022learning} and MGTL \citep{zhao2022mutual} demonstrate significant improvements, particularly in precision metrics like r@1 and r@1\%, where r indicates recall
indicating notable enhancements in model accuracy and robustness. For instance, the MGTL model \citep{zhao2022mutual} achieved the highest scores in r@1 and r@1\% metrics among all listed models on both datasets, signaling a breakthrough in the accuracy of geo-localization systems. These results not only reflect the continual advancements in cross-view geo-localization techniques but also underscore the potential for future applications in diverse fields ranging from automated navigation to augmented reality.

The recall of a classification model is defined as the ratio of true positives (TP) to the sum of true positives and false negatives (FN):

\[
\text{Recall} = \frac{TP}{TP + FN}
\]

This measures the ability of the classifier to find all the positive samples.

\begin{table}[htbp]
\centering
\caption{Comparison of model performance on VIGOR benchmark (Same-area).}
\begin{tabularx}{\textwidth}{|X|X|X|X|X|X|}
\hline
\textbf{Reference paper} & \textbf{R@1} & \textbf{R@5} & \textbf{R@10} & \textbf{R@1\%} & \textbf{Hit Rate} \\
\hline
\citet{shi2019spatial} & 18.69\% & 43.64\% & 55.36\% & 97.55\% & 21.90\% \\
\citet{zhu2021vigor} & 38.02\% & 62.87\% & 71.12\% & 97.63\% & 41.81\% \\
\citet{zhu2021vigor} & 41.07\% & 65.81\% & 74.05\% & 98.37\% & 44.71\% \\
\citet{zhu2022transgeo} & 61.48\% & 87.54\% & 91.88\% & 99.56\% & 73.09\% \\
\citet{zhang2023cross} & 56.51\% & 80.37\% & 86.21\% & 99.25\% & 61.76\% \\
\citet{zhang2023geodtr+} & 59.01\% & 81.77\% & 87.10\% & 99.07\% & 67.41\% \\
\hline
\end{tabularx}
\label{table:comparisonVIGORSameArea}
\end{table}

\begin{table}[htbp]
\centering
\caption{Comparison of model performance on VIGOR benchmark (Cross-area).}
\begin{tabularx}{\textwidth}{|X|X|X|X|X|X|}
\hline
\textbf{Refernce Paper} & \textbf{R@1} & \textbf{R@5} & \textbf{R@10} & \textbf{R@1\%} & \textbf{Hit Rate} \\
\hline
\citet{shi2019spatial} & 2.77\% & 8.61\% & 12.94\% & 62.64\% & 3.16\% \\
\citet{zhu2021vigor} & 9.23\% & 21.12\% & 28.02\% & 77.84\% & 9.92\% \\
\citet{zhu2021vigor} & 11.00\% & 23.56\% & 30.76\% & 80.22\% & 11.64\% \\
\citet{zhu2022transgeo} & 18.99\% & 38.24\% & 46.91\% & 88.94\% & 21.21\% \\
\citet{zhang2023cross} & 30.02\% & 52.67\% & 61.45\% & 94.40\% & 30.19\% \\
\citet{zhang2023geodtr+} & 36.01\% & 59.06\% & 67.22\% & 94.95\% & 39.40\% \\
\hline
\end{tabularx}
\label{table:comparisonVIGORCrossArea}
\end{table}

The evaluation of top models on the VIGOR benchmark presents insightful data on geospatial visual recognition capabilities, distinguishing between same-area and cross-area performance. According to Table \ref{table:comparisonVIGORSameArea}, the TransGeo method by \citet{zhu2022transgeo} outperformed others in the same-area evaluation with an impressive recall at 1 (R@1) of 61.48\%, a significant leap from the 41.07\% reported by the same team's earlier method \citep{zhu2021vigor}. The hit rate, which is a critical indicator of the model's practical applicability, was highest at 73.09\% for TransGeo \citep{zhu2022transgeo}, far surpassing the earlier figures such as the 21.90\% from \citet{shi2019spatial}. Conversely, in the cross-area scenario as presented in Table \ref{table:comparisonVIGORCrossArea}, the GeoDTR+ method by \citet{zhang2023geodtr+} achieves the highest R@1 of 36.01\%, indicating a robust generalization capability when compared to the 2.77\% R@1 by \citet{shi2019spatial}. It is noteworthy that the performance of all models diminishes in cross-area evaluations, yet GeoDTR+ maintains a commendable hit rate of 39.40\%, underscoring its effectiveness in more challenging geospatial tasks.

The comparison of various models in cross-view geo-localization, as depicted in Tables \ref{tab:backbone_comparison} and \ref{table:performance}, highlights the diversity and evolution in this research area. A significant number of models such as CVM-Net \citep{hu2018cvm}, Liu \& Li \citep{Liu_2019_CVPR}, CVFT \citep{shi2020optimal}, SAFA \citep{shi2019spatial}, and \citet{shi2020optimal}, predominantly use VGG16 as their backbone. This indicates the sustained relevance of VGG16 in extracting robust features for geo-localization tasks. While models like SAFA \citep{shi2019spatial} and LPN \citep{wang2021each} demonstrate a balance between computational complexity and parameter efficiency, others like L2LTR \citep{yang2021cross} and EgoTR \citep{yang2021cross} present a higher complexity, possibly aiming for greater accuracy. Innovative approaches are also seen in models like TransGeo \citep{zhu2022transgeo} and the SAIG series \citep{zhu2023simple}, which adopt different architectures, showing a trend towards diversification and specialization in the field.

\section{Applications}
\subsection{Automotive Applications}

\begin{figure}
\begin{center}
\includegraphics[scale=0.13]{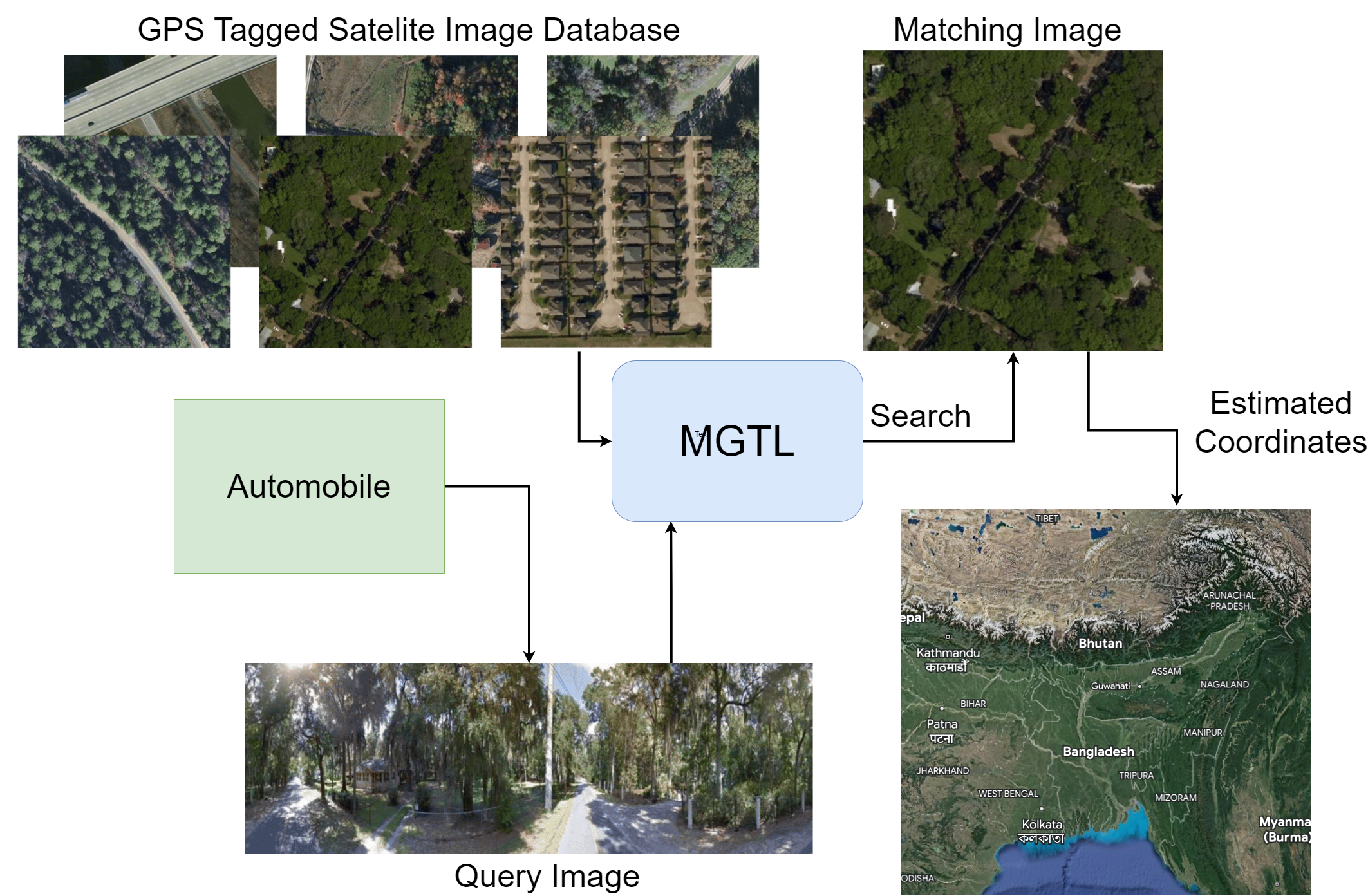}
\caption{Application of algorithm MGTL in automobiles. Figure from \citep{zhao2022co}}\label{fig:MGTL_applications}
\end{center}
\end{figure}

Cross-view geo-localization is a promising field that has immense potential for applications in numerous sectors, including the automotive industry \citep{brosh2019accurate}. This technology seeks to bridge the gap between ground-level and overhead imagery, thereby offering a sophisticated and efficient solution for vehicular geolocation.

Historically, geolocalization has relied heavily on Global Navigation Satellite Systems (GNSS) such as GPS. While these systems are generally reliable, they are susceptible to failures in areas with dense high-rise structures \citep{zhu2018gnss} or in situations where network connectivity is lost. In such scenarios, the reliance solely on GNSS for navigation can lead to inaccuracies and inconsistencies. Cross-view geolocalization offers a robust solution to these problems by providing a means to estimate the vehicle's location using visual cues from ground and aerial views.

A promising application of cross-view geolocalization in the automotive sector is in enhancing traditional navigation systems \citep{shetty2019uav}. For instance, when a vehicle is navigating through densely built urban areas or winding country roads, the onboard navigation system can use this technology to match the live feed from the vehicle's camera with pre-existing satellite imagery. This process allows for more precise location estimation, ensuring more accurate route planning and guidance.

The application of cross-view geolocalization is not limited to standard vehicles. In the world of autonomous vehicles, it has the potential to play a critical role. Self-driving vehicles depend heavily on accurate environmental data for safe and efficient navigation. By incorporating cross-view geolocalization into their navigation systems, these vehicles can better comprehend their surroundings and make more informed decisions. This technology can help overcome challenges posed by complex urban environments or varying weather and lighting conditions, significantly improving autonomous navigation.

The integration of cross-view geolocalization (CVG) into Advanced Driver Assistance Systems (ADAS) could significantly enhance the latter's effectiveness in augmenting specific driving functions and thereby improve the safety and convenience of the driving experience \citep{wang2023satellite}. Firstly, CVG, by virtue of matching ground-level images with aerial or satellite images, facilitates improved localization, a fundamental requirement for the effective functioning of ADAS. Secondly, in scenarios where Global Navigation Satellite System (GNSS) signals are unreliable or obstructed \citep{ghanem2023leveraging}, CVG can provide a viable alternative for precise vehicle localization, which is critical for the safety features of ADAS. Such enhanced localization capabilities could in turn enable ADAS to provide more accurate and reliable assistance in key driving functions such as parking, collision detection and avoidance, and lane departure warnings. Moreover, by continuously matching real-time ground-level imagery with aerial imagery, CVG could potentially help ADAS-equipped vehicles adapt to dynamic environmental changes, ensuring a smoother and safer driving experience. 

Additionally, the advent of the VIGOR \citep{zhu2021vigor} benchmark represents a significant stride in practical cross-view geolocalization. The traditional approach assumes that each ground-level image matches with a single overhead image, an assumption that might not hold true in real-world scenarios. VIGOR, in contrast, recognizes that each ground-level image could match multiple overhead images and vice versa, which is more reflective of real-world conditions. This understanding could be crucial for several automotive applications.

For instance, consider an autonomous vehicle navigating an unfamiliar city. With a traditional one-to-one matching system, the vehicle might struggle to accurately determine its location if the ground-level view doesn't perfectly match any single overhead image. With a system like VIGOR, however, the vehicle could use multiple overhead images to better estimate its location, making for safer and more efficient navigation.

Similarly, in situations where the road layout has changed since the overhead images were taken - due to construction, for instance - a traditional system might struggle. But a system based on the VIGOR benchmark could use the multiple matches to piece together an accurate picture of the current layout and adjust its navigation accordingly.

The results summarized in Table ~\ref{tab:my_label} present a comparative performance analysis. EgoTR \cite{yang2021cross} being the first implementation of a transformer showcase higher computational demands, as indicated by their higher GFLOPs and parameter count, which might limit their real-time deployment in resource-constrained environments. Interestingly, the TransGeo \citep{zhu2022transgeo} which makes use of attention based cropping illustrate a significant reduction in GFLOPs while maintaining a moderate parameter size.  Models like SAFA \citet{shi2019spatial} and \citet{shi2020optimal} demonstrate a balance between GFLOPs and parameters, indicating optimized processing capabilities. Particularly, SAFA, with 40.2 GFLOPs and 29.5 million parameters, achieves a notable frame rate of 35.90 FPS, highlighting its potential for real-time applications. These advancements in cross-view geo-localization models are crucial in enhancing the accuracy and speed of location estimation in dynamic environments, thereby augmenting the capabilities of autonomous navigation systems in real-world scenarios.

\begin{table}[htbp]
\centering
\caption{Comparison of FLOPs, parameters, and computational speed in cross-view geo-localization.}
\label{tab:my_label}
\begin{tabularx}{\textwidth}{|X|X|X|X|X|}
\hline
\textbf{Model} & \textbf{Backbone} & \textbf{GFLOPs} & \textbf{\#Params (M)} & \textbf{\#FPS} \\
\hline
CVM-Net \citep{hu2018cvm} & VGG16 & - & 160.3 & - \\
\hline
CVFT \citep{shi2020optimal} & VGG16 & - & 26.8 & - \\
\hline
SAFA \citep{shi2019spatial} & VGG16 & 40.2 & 29.5 & 35.90 \\
\hline
\citet{shi2020optimal} & VGG16 & 39.3 & 14.7 & 22.72 \\
\hline
LPN \citep{wang2021each} & VGG16 & 40.2 & 29.5 & - \\
\hline
L2LTR \citep{yang2021cross} & HybridViT & 57.1 & 195.9 & - \\
\hline
TransGeo \citep{zhu2022transgeo} & DeiT-S/16 & 12.3 & 44.9 & - \\
\hline
SAIG-S \citep{zhu2023simple} & SAIG-S & 8.8 & 18.2 & - \\
\hline
SAIG-D \citep{zhu2023simple} & SAIG-D & 13.3 & 31.2 & - \\
\hline
EgoTR \citep{yang2021cross} & ResNet \& ViT & - & 195.9 & 13.37 \\
\hline
TransGCNN \citep{wang2022transformer} & VGG16 \& ViT & - & 87.8 & 23.09 \\
\hline
\end{tabularx}
\end{table}

However, the implementation of cross-view geolocalization in the automotive industry comes with certain challenges. The requirement for large amounts of data for matching ground and aerial views, the necessity for real-time processing for practical applications, and the high degree of accuracy needed for safe navigation are all substantial hurdles. Moreover, changes in the environment due to construction, natural disasters, or seasonal variations can further complicate the geolocalization process.

\subsection{Robotics Applications}
Cross-view geolocalization holds immense potential for a variety of robotic applications. In numerous scenarios, robots operate in environments where GPS signals are unreliable or entirely blocked. These scenarios could include urban settings with high buildings that create "urban canyons" or indoor environments where GPS signals cannot penetrate. In such situations, cross-view geolocalization, such as the models Wide-Area Geolocalization (WAG) and Restricted FOV Wide-Area Geolocalization (ReWAG) \citep{downes2022wide}, offer robust solutions for accurate, real-time localization.

For instance, autonomous ground robots navigating in urban or indoor environments can leverage ReWAG \citep{downes2022wide} for localization. Given that these robots typically carry standard cameras (as opposed to panoramic cameras), the pose-aware embeddings and particle pose incorporation strategies of ReWAG become valuable for accurate geolocation. By matching the images captured by the robots with corresponding satellite or aerial-view images, the robots can determine their positions even in the absence of reliable GPS signals.

Apart from navigation, cross-view geolocalization can aid in various other robotic applications such as environmental monitoring and mapping \citep{biswas2011depth}. Robots equipped with cameras can capture ground-level images, and through cross-view geolocalization, these images can be mapped onto corresponding overhead images. This can help in creating detailed, high-resolution maps of the environment \citep{xia2021cross}, useful in numerous domains like urban planning, disaster management, and environmental studies.

\subsection{Augmented Reality}
Augmented Reality (AR) has emerged as a potent technology in a myriad of applications, leveraging the power of superimposing digital content onto the real world. One of the crucial elements in creating immersive AR experiences is precise geo-localization. The integration of cross-view geo-localization into AR applications \citep{mithun2023cross} has transformative potential, enhancing both user experience and spatial understanding.

Cross-view geo-localization, where a ground-level image is matched against geotagged aerial or satellite images to determine its geographic location, plays a vital role in AR. By accurately determining the location and orientation of a camera relative to a 3D scene, AR applications can seamlessly overlay digital content onto the physical world with a high degree of precision. In particular, a robust cross-view geo-localization can improve the visual alignment of the digital and physical objects \citep{wilson2021visual}, thereby increasing the immersion and interactivity of AR experiences.

AR navigation \citep{emmaneel2023cross} is a compelling application where cross-view geo-localization can have a transformative impact. When navigating complex urban environments, AR can provide intuitive, contextually-aware directions overlaid onto the real world, creating a more immersive and user-friendly experience. To achieve this, precise geo-localization is critical. Using cross-view methods, a user's ground-level view can be accurately matched with aerial or satellite imagery, allowing for precise location tracking and enhancing the reliability of the overlaid navigational cues.

Moreover, AR tourism \citep{shih2019arts} can also benefit greatly from cross-view geo-localization. Tourists exploring a city with an AR application can have information about nearby landmarks or points of interest superimposed on their live camera feed. By using cross-view geo-localization, these applications can accurately determine the user's location and orientation, ensuring that the digital content aligns correctly with the real-world landmarks.

However, implementing cross-view geo-localization in AR poses significant challenges. Resource constraints on mobile devices can hinder the performance of localization methods that rely on memory-intensive local descriptors \citep{middelberg2014scalable}. This has led to the exploration of scalable methods, combining the robustness of global localization systems with the speed and efficiency of local pose tracking on mobile devices. Solutions like these could significantly enhance AR applications by providing accurate, real-time geo-localization in a scalable manner.

Cross-view geo-localization holds immense promise for enhancing AR applications. By improving the accuracy of location and orientation tracking, it can create more immersive, interactive, and contextually-aware AR experiences. As research continues in this field, we can expect to see more sophisticated and reliable AR applications that fully exploit the potential of cross-view geo-localization.

\subsection{Drone Navigation and UAV localization}
UAV localization and pose estimation \citep{ding2020practical, shetty2019uav, tian2021uav, kan2022target, bui2022part, luo2022deep} is a topic close to cross view geolocalization and lends itself to knowledge transfer between the two topics. In traditional setups, drones rely heavily on GPS for navigation. However, in certain scenarios like dense urban environments or inside large structures, GPS signals can be unreliable or entirely inaccessible. Here, cross-view geo-localization becomes particularly useful. This technique leverages the ability to match ground-level imagery captured by the drone with pre-existing satellite or aerial imagery to determine the precise location of the drone.

Drones equipped with high-resolution cameras can capture detailed imagery of the terrain below. These images can then be compared and matched with pre-existing satellite images using deep learning techniques \citep{zheng2020university}. However, this task is complicated by the significant differences in perspective, scale, and details between ground-level and aerial/satellite images. Recent advances in deep learning algorithms have allowed for the development of models that can overcome these challenges, creating robust mappings between ground-level and aerial views.

Further, drones operating in complex environments often have to deal with dynamic elements, such as moving objects or changing lighting conditions. Advanced cross-view geo-localization techniques can help drones navigate these environments by providing them with up-to-date location information, facilitating autonomous navigation and decision-making \citep{ding2020practical}. In addition to providing positional data, these techniques can also offer contextual information about the environment, aiding in path planning and obstacle avoidance.

\citet{middelberg2014scalable} pointed out how to achieve accurate, real-time geo-localization on mobile devices. In terms of cross-view geo-localization, the 6-DOF (six degrees of freedom) localization capability ensures that the system can determine both the position (in three dimensions) and the orientation (rotation about three axes) of the camera relative to the scene. This could be particularly beneficial in applications involving complex urban environments where changes in both position and orientation of the camera can occur frequently.

\section{Discussion and Future}
The application of advanced deep learning techniques, particularly those based on transformer architectures, such as BERT (Bidirectional Encoder Representations from Transformers) or more recent models like BEiT (BERT for Image Transformations), has the potential to significantly impact the field of cross-view geolocalization. The power of transformers in managing sequence data, originally showcased in Natural Language Processing (NLP), has proven to be translatable to image data and promises an innovative approach to cross-view geolocalization.

Transformers are designed to handle the dependencies between input and output without regard to their distance in the input or output sequences. This property is particularly beneficial for tasks that require understanding and encoding contextual information, such as cross-view geolocalization. As such, the adoption of transformer-based architectures can provide a more robust and efficient way to handle the task's inherent complexities. As it is being seen already there has been a move towards transformer based architecture already in this topic, for example, TransGeo model \citep{zhu2022transgeo}. However, this is a rapidly advancing field and as such much more improved and specialized models can be built using transformers.

 ViT \citep{dosovitskiy2010image} applies the transformer architecture directly to sequences of image patches and has shown competitive results against traditional convolutional architectures on large-scale image recognition tasks. Its ability to handle global relationships between patches in an image could be exploited in cross-view geolocalization, allowing the model to better understand the overall structure of the scene and improve the accuracy of localization.

The adoption of BEiT \citep{bao2021beit} specifically can offer even more benefits. As an image-based transformer model, BEiT has been pre-trained using a large amount of image data and thus, has learned to capture rich visual features and structures that can be vital for geolocalization tasks. By leveraging BEiT, we could potentially reduce the amount of training data needed and the training time required for our cross-view geolocalization model. Furthermore, the model could deliver improved performance by exploiting the high-level visual features learned by BEiT.

However, it's important to note that while transformer-based models offer significant potential, applying them in the context of cross-view geolocalization is not without challenges. Transformers require significant computational resources and large amounts of training data, which can be limiting factors in their adoption. There are also challenges related to handling the multi-modal nature of cross-view geolocalization data, and the scale of the task, which involves dealing with high-resolution satellite and ground-view images covering large geographical areas.

The use of advanced techniques such as domain adaptation and multi-task learning, alongside transformer-based models, could potentially address these issues. Domain adaptation techniques can enable the model to better transfer learned features from the source (satellite-view images) to the target domain (ground-view images). Multi-task learning, on the other hand, can enable the model to learn more generalized features by learning to perform multiple related tasks simultaneously.

Deformable convolutions, an innovation in convolutional neural network (CNN) architecture, offer a significant leap in handling geometric transformations in images. Traditional CNNs struggle with variations in scale, viewpoint, and spatial layout because of their fixed geometric structure. However, deformable convolutions address this by introducing modifiable or `deformable' convolutional kernels that can adapt to the spatial transformations in the input data. In the context of cross-view geo-localization, this property is particularly beneficial. The images from different views, such as ground-level and aerial views, often have significant variations in viewpoint and spatial layout. By applying deformable convolutions, the model can adjust its kernels according to the specific spatial transformations between the two views. This results in more accurate feature extraction and thus, better cross-view matching.

There is substantial potential in leveraging advanced deep learning techniques such as BEiT and other transformer models in cross-view geo localization. While challenges persist, the potential benefits of improved performance, reduced training time, and data efficiency make these techniques promising directions for future research in this field. It will be fascinating to see how these methods will continue to evolve and influence the landscape of cross-view geolocalization in the years to come.


\section{Conclusion}
Cross-view geo-localization is a rapidly evolving domain, pivoting from foundational pixel-wise methods to sophisticated deep learning techniques. This paper delivered an exhaustive review of geo-localization strategies, categorizing them from the lens of their evolution: pixel-wise approaches, feature-based techniques, and revolutionary contributions from models like Siamese networks, capsule networks, and attention-based mechanisms.

While the focus was largely on recent advancements, it is paramount to approach future challenges armed with knowledge of prior milestones. In tracing the history and current state of geo-localization, we have developed a framework that facilitates an understanding of contemporary research and pinpoints areas warranting deeper investigation. Emphasis was placed on applications like orientation context, attention-based models, and benchmark datasets like CVUSA and CVACT.

Within geo-localization, the potential of pre-training strategies, especially when combined with transformers and innovative techniques like deformable attention, is immense. Transformers can grasp intricate spatial relations, while deformable attention adeptly adapts to diverse spatial patterns inherent in cross-view data. These methods promise to tackle challenges like viewpoint variations and improve localization accuracy. Through our review, we aim to inspire and pave the way for groundbreaking research in these promising avenues.

\bibliography{manuscript}

\end{document}